\def\year{2021}\relax
\documentclass[letterpaper]{article} 
\usepackage{aaai21}  
\usepackage{times}  
\usepackage{helvet} 
\usepackage{courier}  
\usepackage[hyphens]{url}  
\usepackage{graphicx} 
\usepackage{graphicx}  
\usepackage{natbib}  
\usepackage{caption} 
\usepackage{amsfonts}       
\usepackage{amsmath}
\usepackage{booktabs} 
\usepackage{multirow} 
\usepackage{subcaption}

\title{Memory-Gated Recurrent Networks}
\author {
    Yaquan Zhang\textsuperscript{\rm 3} \quad
    Qi Wu\textsuperscript{\rm 2}\footnote{Corresponding author.}\quad
    Nanbo Peng\textsuperscript{\rm 1}\quad 
    Min Dai\textsuperscript{\rm 4}\quad 
    Jing Zhang\textsuperscript{\rm 2} \quad Hu Wang\textsuperscript{\rm 1}\\
}

\affiliations {
    \textsuperscript{\rm 1} JD Digits, pengnanbo@jd.com, wanghu5@jd.com\\
    \textsuperscript{\rm 2} City University of Hong Kong, qiwu55@cityu.edu.hk,  jingzha28-c@my.cityu.edu.hk \\
    \textsuperscript{\rm 3} National University of Singapore, Department of Mathematics and Risk Management Institute, rmizhya@nus.edu.sg\\
    \textsuperscript{\rm 4}National University of Singapore,  Department of Mathematics, Risk Management Institute, and Chong-Qing \& Suzhou
    Research Institutes, mindai@nus.edu.sg\\
}
\begin{document}
\maketitle
\begin{abstract}
The essence of multivariate sequential learning is all about how to extract dependencies in data. These data sets, such as hourly medical records in intensive care units and multi-frequency phonetic time series, often time exhibit not only strong serial dependencies in the individual components (the ``marginal" memory) but also non-negligible memories in the cross-sectional dependencies (the ``joint" memory). Because of the multivariate complexity in the evolution of the joint distribution that underlies the data generating process, we take a data-driven approach and construct a novel recurrent network architecture, termed Memory-Gated Recurrent Networks (mGRN), with gates explicitly regulating two distinct types of memories: the marginal memory and the joint memory. Through a combination of comprehensive simulation studies and empirical experiments on a range of public datasets, we show that our proposed mGRN architecture consistently outperforms state-of-the-art architectures targeting multivariate time series.  
\end{abstract}

\section{Introduction}
A multivariate time series consists of values of several time-dependent variables. The data structure commonly appears in fields such as economics and engineering. By studying multivariate time series, one can make forecasts based on past observations or determine which category the time series belongs. Despite its importance, multivariate time series analysis is a daunting task due to the complexity of the data structure. The values of each variable may not only depend on the past self but also interact with other variables. 

Machine learning algorithms are commonly adopted in the analysis of multivariate time series. Early attempts can be traced back to  \citet{Chakraborty1992}, in which feedforward neural networks are studied. Later on, a specialized architecture known as recurrent neural networks (RNN) was proposed. They are specially designed to handle sequential inputs. Among the variants of RNN, long short-term memory (LSTM) units \citep{Hochreiter1997} and gated recurrent units (GRU)  \citep{Cho2014} are perhaps the most popular. They introduce intermediary variables, which are referred to as gates, to regulate memory and information flow. More recently, some advanced algorithms were proposed. State-of-the-art results were obtained by WEASEL+MUSE \citep{Schaefer2017},  MLSTM-FCN \citep{Karim2019}, channel-wise LSTM \citep{Harutyunyan2019}, and TapNet \citep{Zhang2020}, just to name a few.

As alternatives to machine learning algorithms, one may also analyze multivariate time series with traditional time series models. An example is the multivariate ARMA-GARCH model \citep{Ling2003}, in which one models the evolution of each variable with an ARMA process and models the covariance of variables by a GARCH process. The drawback of this approach is that it makes strict assumptions on data structures such as innovation distributions and linearity of dependence. The assumptions are often not flexible enough to deal with real-world data sets. 

Despite their drawbacks, the traditional time series models have an important implication: \textit{the dynamics of multivariate time series can be separately described by the individual dynamic of each variable (the ARMA processes) and the dynamic of variable interactions (the GRACH process).} The seemingly trivial implication motivates us to propose  Memory-Gated Recurrent Network (mGRN)\footnote{https://github.com/yaquanzhang/mGRN}, which modifies the existing RNN architectures to match the multivariate time series's internal structure. Specifically, we split variables into a few groups. For each variable group, we set up a marginal-memory component regulating only the group-specific memory and information. Afterward,  candidate memories of marginal components are combined and regulated in the joint-memory component to learn interactions among variable groups. In this way, we establish the correspondence between memory and information within each variable group. Such an explicit correspondence is missing in the existing RNN architectures. 

To demonstrate the superiority of mGRN, we extensively test the model with simulated and real-world data sets. In the simulation study, the multivariate time series data are generated from the heavy-tailed model proposed in \citet{Yan2019}. We design a task such that a good prediction precisely requires the learning of both idiosyncratic and joint serial dependence. The real-world data sets are borrowed from recent state-of-the-art papers \citet{Harutyunyan2019} and \citet{Zhang2020}. The data sets are collected from a wide range of applications, such as hourly medical records in intensive care units and multi-frequency phonetic time series. Compared with the strong baselines established in the literature, the proposed model makes significant and consistent improvements.

The rest of the paper is organized as follows. In Section \ref{sec:related_work}, we briefly review popular algorithms targeting multivariate time series. The architecture of the proposed mGRN is presented in Section \ref{sec:architeture}. The experiments with simulated and empirical data sets are respectively provided in Section \ref{sec:simulation} and \ref{sec:realdatasets}. Lastly, we conclude in Section \ref{sec:conclusion}.   

\section{Related Work}
\label{sec:related_work}
In this section, we briefly review some widely adopted methods to study multivariate time series. Generally speaking, the available methods can be grouped into three categories.


The first category is based on classical machine learning techniques. Dynamic time warping (DTW) classifies univariate time series by measuring the distances among samples. To handle the multivariate cases, there are a few popular variants, namely ED-I, DTW-I, and DTW-D; see \citet{Shokoohi-Yekta2015} for a review. An alternative approach is known as word extraction for time series classification with multivariate unsupervised symbols and derivatives (WEASEL+MUSE) \citep{Schaefer2017}. The algorithm is based on the bag-of-words framework and extracts features by Chi-square tests. WEASEL+MUSE is shown to outperform similar algorithms such as gRSF \citep{Karlsson2016}, LPS \citep{Baydogan2016}, mv-ARF \citep{Tuncel2018} and SMTS \citep{Baydogan2015}.  However, WEASEL+MUSE may cause memory issues when the underlying data set is large \citep{Zhang2020}.

The second category involves neural networks. Although vanilla LSTM has been proposed for more than two decades, it is still arguably the most popular neural network in multivariate time series analysis. LSTM has achieved state-of-the-art results in many applications; see \citet{Lipton2015} for a detailed review. More recently, \citet{Tai2015} proposes tree LSTM for natural language processing tasks. It modifies the sequential information propagation in vanilla LSTM to a tree-structured network to capture non-sequential dependence among words. Dipole \citep{ma2017} combines bidirectional GRU and the attention mechanism to study multivariate clinical data sets. LSTM-FCN \citep{Karim2017} combines LSTM and convolutional neural networks (CNN) to handle univariate time series. \citet{Karim2019} proposes MLSTM-FCN, which extends LSTM-FCN with a squeeze-and-excitation block to handle the multivariate cases. \citet{Zhang2020} proposes an attentional prototype network known as TapNet. It makes use of feature permutation and CNN to extract low-dimensional feature representations from multivariate data.

There have been previous attempts to separately handle marginal and joint memories of multivariate time series within this category. \citet{Chakraborty1992} compares a joint feed-forward network for all variables with separate networks for each variable. The conclusion is that the joint network works better, which is indeed expected. Unlike the proposed model, the separate networks cannot capture the interactions among variables. More recently, \citet{Belletti2018} proposes block-diagonal RNN for natural language processing tasks. The architecture splits variables into a few groups and applies an RNN to each group. The motivation is to learn long-range memory by introducing large gates while reducing the computation burden. However, note that the information from each block is combined by merely a fully connected layer, so the memory in the interactions among variable groups is lost. \citet{Harutyunyan2019} proposes channel-wise LSTM, in which there is an LSTM layer for each variable, and the outputs are concatenated and fed into another LSTM layer. The architecture is demonstrated to outperform vanilla LSTM.  Although channel-wise LSTM shares a similar intuition with mGRN, we demonstrate that mGRN benefits from the deliberately designed gates and information flows and possesses clear advantages; see more discussions in Section \ref{sec:architeture} and experiments in Section \ref{sec:simulation} and \ref{sec:mimic_3}. 

The last category is the traditional statistical time series models. Despite the fact that machine learning techniques are prevailing, traditional statistical models are still widely applied to time series analysis in fields such as economics and finance \citep{Tsay2005}. Compared with machine learning techniques, statistical models are easy to implement and explain. Two primary tools in this category are VARMA \citep{Quenouille1957} and ARMA-GARCH \citep{Ling2003}. VARMA models the multivariate dependence by linear ARMA processes. ARMA-GARCH models the evolution of each variable with a linear ARMA process, and the covariance of variables with a GARCH process.

\section{Memory-Gated Recurrent Networks}
\label{sec:architeture}
As GRU serves as a building block of mGRN, we begin with a review of its structure. It simplifies the gates and memory flows of LSTM. 

Suppose we are given a multivariate time series $X_t$ containing $M$ variables. At step $t$, the inputs of a GRU with unit dimension $N$ include the sequential data input $X_t\in \mathbb{R}^{M\times 1}$, and the output $h_{t-1}\in \mathbb{R}^{N\times 1}$ from the previous step. The inputs are regulated by the reset gate $r\in\mathbb{R}^{N\times\ 1}$ and the update gate $z\in\mathbb{R}^{N\times 1}$. 
\begin{equation*}
\begin{aligned}
r_t = \sigma(W_r X_{t} + U_r h_{t-1} + b_r), \\
z_t = \sigma(W_z X_{t}+ U_z h_{t-1} + b_z),\\
\end{aligned}
\end{equation*}
where $\sigma(\cdot)$ is a sigmoid function. The reset gate $r$ is used to generate the candidate memory $\tilde{h}$ by incorporating the new information $X_t$. $\tilde{h}$ is then used to construct the final memory output $h$ together with the update gate $z$.
\begin{equation*}
\begin{aligned}
&\tilde{h}_t = \tanh(W_h X_{t} + r_t\odot(U_h h_{t-1}) + b_h),\\
&h_t = (1-z_t)\odot h_{t-1} + z_t\odot \tilde{h}_t,
\end{aligned}
\end{equation*}
where $\odot$ stands for the element-wise product of two vectors. $W_r$, $W_z$, $W_h\in \mathbb{R}^{N\times M}$, $U_r$, $U_z$, $U_h\in \mathbb{R}^{N\times N}$, and 
$b_r$, $b_z$, $b_h\in \mathbb{R}^{N\times 1}$ are the trainable parameters.

\begin{figure*}
	\centering
	\includegraphics[width=0.85\linewidth]{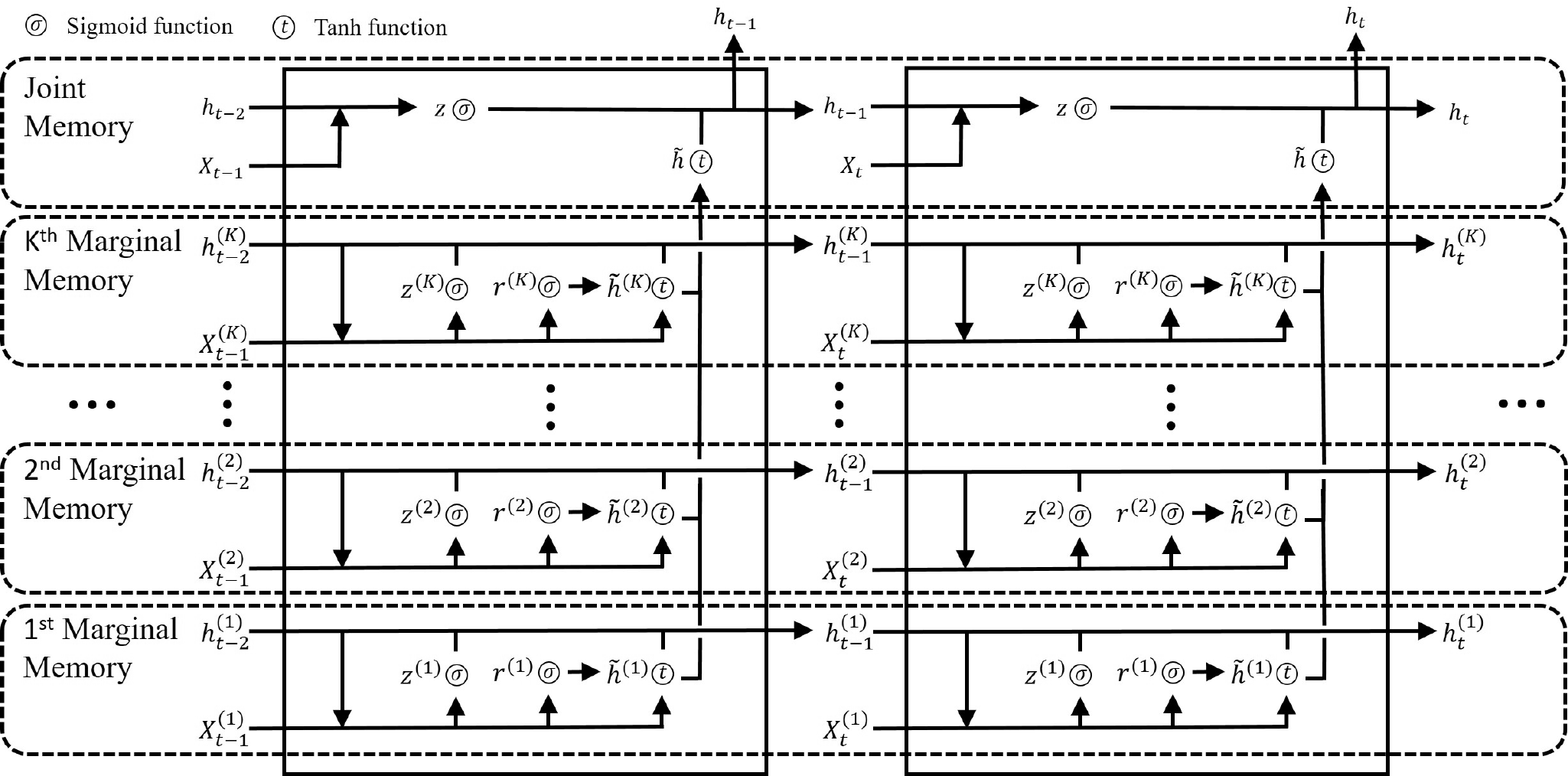}
	\caption{Illustration of mGRN. }
	\label{fg:illustration}
\end{figure*}

As implied by traditional statistical time series models, the multivariate time series's internal structure can be split to self-dependent components of each variable group, and the interactive component among variables. However, in a vanilla LSTM or GRU, it relies on the neural network to figure out the internal structure. In mGRN, we set up marginal-memory components to extract the group-specific information and a joint-memory component for the joint information, respectively. By explicitly setting up the components, the task of neural networks is simplified. 

The mGRN architecture is as follows. An illustration is given in Figure \ref{fg:illustration}.

\textbf{The marginal-memory component}. Suppose the multivariate time series $X_t$ is divided into $K$ groups, i.e. $X_t = [X^{(1)}_t,\dots,X^{(K)}_t]$, where each $X^{(k)}_t$ consists of $m_k$ variables with $\sum_{k=1}^K m_k = M$. Marginal-memory components are formulated in the form of GRU. Given the component dimension $N_k$, the $k$-th marginal component contains the candidate memory 
$\tilde{h}^{(k)}\in \mathbb{R}^{N_k\times 1}$ and the final memory $h^{(k)}\in \mathbb{R}^{N_k\times 1}$, which are controlled by their own reset gate $r^{(k)}\in\mathbb{R}^{N_k\times 1}$ and update gate $z^{(k)}\in\mathbb{R}^{N_k\times 1}$. 
\begin{equation*}
\begin{aligned}
&r^{(k)}_t = \sigma\left(W^{(k)}_r X^{(k)}_t + U^{(k)}_r h^{(k)}_{t-1} + b^{(k)}_r\right),\\
&z^{(k)}_t = \sigma\left(W^{(k)}_z X^{(k)}_t + U^{(k)}_z h^{(k)}_{t-1} + b^{(k)}_z\right),\\
&\tilde{h}^{(k)}_t = \tanh\left(W^{(k)}_h X^{(k)}_t + r^{(k)}_h\odot \left(U^{(k)}_h h^{(k)}_{t-1}\right) + b^{(k)}_h\right),\\
&h^{(k)}_t = \left(1-z^{(k)}_t\right)\odot h^{(k)}_{t-1} + z^{(k)}_t\odot \tilde{h}^{(k)}_t.
\end{aligned}
\end{equation*}
The trainable parameters are $W^{(k)}_r$, $W^{(k)}_z$, $W^{(k)}_h\in \mathbb{R}^{N_k\times M_k}$, $U^{(k)}_r$, $U^{(k)}_z$, $U^{(k)}_h\in \mathbb{R}^{N_k\times N_k}$, and 
$b^{(k)}_r$, $b^{(k)}_z$, $b^{(k)}_h\in \mathbb{R}^{N_k\times 1}$. 

\textbf{The joint-memory component.} The joint-memory component is constructed as follows. Given the component dimension $N$, we combine the marginal candidate memories $\{\tilde{h}^{(k)}\}_{k=1}^K$ to form the joint candidate memory $\tilde{h}\in \mathbb{R}^{N\times 1}$ as
\begin{equation*}
\tilde{h}_{t} = \tanh\left(\sum_{k=1}^K U^{(k)}_c \tilde{h}^{(k)}_t + b_c\right).
\end{equation*}
We then add a joint update gate $z\in \mathbb{R}^{N\times 1}$ to regulate the final full output memory $h$ as
\begin{equation*}
\begin{aligned}
&z_t = \sigma(W_z X_{t}+ U_z h_{t-1} + b_z), \\
&h_t = (1-z_t)\odot h_{t-1} + z_t\odot \tilde{h}_t, \\
\end{aligned}
\end{equation*}
In this component, the trainable parameters are $W_z\in \mathbb{R}^{N\times M}$, $ U^{(k)}_c\in \mathbb{R}^{N_k\times N}$, $ U_z\in \mathbb{R}^{N\times N}$ and $b_c$, $b_z\in\mathbb{R}^{N\times 1}$.

There are a few remarks regrading the mGRN architecture. 
\begin{enumerate}
	\item The marginal-memory components only involve group-specific inputs and memories. The network does not need to pick up input-memory correspondence from the mixed data as in vanilla LSTM or GRU.
	
	\item We expose the candidate marginal memories $\{\tilde{h}^{(k)}\}_{k=1}^K$, instead of the final marginal memories $\{h^{(k)}\}_{k=1}^K$, to the joint component. Our intuition is to give the joint component only the new information. The joint component has its own memory and updates gate to determine which part of the new information is useful.
	
	\item Compared with vanilla GRU and LSTM, separately handling marginal and joint memories will inevitably lead to more intermediate gates. To avoid unnecessarily large models, we take a conservative approach in the design of mGRN. Instead of LSTM, we choose GRU as the building block since GRU achieves comparable performance \citep{Chung2014} with fewer gates. Instead of using a GRU layer as the joint component, we pick up gates that are crucial to the performance through preliminary experiments. As a result, given the same number of trainable parameters, gates in mGRN have greater sizes comparing with channel-wise LSTM. It is interesting to note that the simplified gates improve model performance not only in the case that the total number of trainable parameters is controlled (see Section \ref{sec:simulation}), but also in the case that all hyperparameters are free to tune.
	
	\item Channel-wise LSTM artificially doubles the features by reversing the input time series. We purposely exclude the step in mGRN to avoid the confusion that the improvements of mGRN may come from reversing the inputs. The technique can be applied to augment any algorithms. In fact, mGRN outperforms channel-wise LSTM even without the extra step; see Section \ref{sec:simulation} and \ref{sec:mimic_3}.

	\item In the traditional time series models such as ARMA, self-dependence is usually limited to a single variable. In mGRN, we extend the scope of self-dependence to a variable group. The internal structures of data sets suggest the extension. For example, in the MIMIC-III data sets in Section \ref{sec:mimic_3}, each variable is coupled with a binary variable to indicate whether it is observed at the step or not. It is intuitive to include the primary variable and the indicative variable in the same group. In the case that the domain knowledge is inadequate to determine variable grouping, we recommend starting from a total split of all variables, which usually gives a satisfactory performance in our experiments. In practice, variable grouping can be regarded as a step of hyperparameter tuning. 
	
	\item Following the settings of channel-wise LSTM, in all experiments, the dimensions of marginal components are chosen to be the same, i.e. we choose $\tilde{N}$ such that $N_k = \tilde{N}$ for all $k$. Moreover, the joint memory dimension is chosen relative to $\tilde{N}$. We tune a variable $\lambda$ with $N = \lambda\tilde{N}$. In most experiment cases, mGRN performs the best with $\lambda = 2$ or $\lambda = 4$, i.e. the joint component should be much greater than marginal components.  
	
	\item In this paper, we purposely keep the architecture simple to emphasize on the impressive improvements brought by learning marginal and joint memories separately. The architecture can be readily augmented with other components such as CNN and the attention mechanism. To explore the combinations so that the model performance can be maximized will be our future research.

\end{enumerate}

\section{Simulation Experiments}
\label{sec:simulation}
In this section, we present a prediction task based on simulated time series. The task is specially designed such that a good prediction requires learning of both marginal and joint serial dependence. Despite both channel-wise LSTM and mGRN are designed for this task, mGRN demonstrates consistent improvements.
\subsection{Data Generation Process}
The data generation process is taken from \citet{Yan2019}, but with modifications to allow both idiosyncratic and joint serial dependence. To be specific, we generate two correlated series given by
\begin{equation}
\label{eq:simulation_y}
\begin{aligned}
y_i(t) = &\alpha_i(t) + \beta_i(t)g(\omega_{M}(t); u_{M, i}(t), v_{M, i}(t)) \\
&+ \gamma_i(t)g(\omega_i(t); u_i(t), v_i(t)), \quad \text{for} \quad i = 1, 2,
\end{aligned}
\end{equation}
where $g(\omega;u, v) := \omega(u^\omega/A + v^{-\omega}/A + 1)$ with $A = 4$ and $u\geq0$, $v\geq0$\footnote{In \citet{Yan2019}, function $g(\omega; u, v)$ is defined with $u$, $v \geq 1$. The requirement is imposed to make sure the distribution is heavy tail, which is commonly observed in financial data. We relax the requirement for the simplicity of data generation. }. $\omega_{M}(t)$ and $\omega_i(t)$ are independent with common distribution $ \mathcal{N}(0,1)$. 

The parameters have serial-dependent dynamics given by the following AR(5) process
\begin{equation}
\label{eq:simulation_ar}
\begin{aligned}
p(t) =& \mu_{p} + 0.9p(t-1) - 0.8p(t-2) + 0.7p(t-3) - \\
&0.6p(t-4) + 0.5p(t-5) + \epsilon_{p}(t) \\
\text{for }  p = & \alpha_i, \log{\beta_i}, \log{u_{M, i}}, \log{v_{M, i}}, \log{\gamma_i}, \log{u_i}, \log{v_i}, \\ 
 i = & 1, 2,
\end{aligned}
\end{equation}
where the random noises $\epsilon_{p}(t)$ are independent with common distribution $\mathcal{N}(0,0.01)$. Note that we take the logarithm of parameters other than $\alpha$ to ensure positivity. 

The coefficients of the AR processes (\ref{eq:simulation_ar}) are arbitrarily fixed except for the constant terms, which are chosen to generate realistic time series. Table 2 of \citet{Yan2019} reports the parameters of a few stocks. $y_1$ and $y_2$ are matched with a pair of stocks in the table. $\mu_{\alpha_i}$ is selected so that $E[\alpha_i(t)]$ matches with the corresponding parameter. For $p = \log{\beta_i}$ and $\log{\gamma_i}$, $\mu_{p}$ is chosen such that $\exp{E[p(t)]}$ matches with one tenth\footnote{The purpose is to enhance predictability.} of the corresponding parameter. For $p =  \log{u_{M, i}}, \log{v_{M, i}}, \log{u_i}$ and $\log{v_i}$, $\mu_{p}$ is chosen such that $\exp{E[p(t)]}$ matches with the corresponding parameter. To guarantee the robustness of the results, we repeat experiments with 10 pairs of randomly selected stocks\footnote{The list of randomly selected stock pairs, together with the corresponding constant terms, are given in Appendices.}.

The simulation process (\ref{eq:simulation_y}) is proposed initially to capture the extremal dependence among financial assets. We adopt this model in the simulation study for a few reasons. Firstly, it explicitly sets up an idiosyncratic component and a correlated component, which match with the structures of mGRN. Moreover, it provides the simulated data with adequate complexity, such as randomness, heavy-tailless and time-varying distributions, all of which are commonly observed in real-world data sets. 

\subsection{Analysis of the Prediction Task}
The task is to jointly predict $100\times y_1(t)\times y_2(t)$ given observations up to step $t-1$. We choose the loss function to be mean squared error (MSE), so that there is a theoretical best predictor $E[100y_1(t)y_2(t)|\mathcal{F}_{t-1}]$ \citep{Shumway2017}. Thanks to the Gaussian conditional distributions of AR processes (\ref{eq:simulation_ar}), the best predictor can be explicitly evaluated. 

We postpone the tedious evaluation of the best predictor $E[100y_1(t)y_2(t)|\mathcal{F}_{t-1}]$ to Appendices, and focus on the implications. First, the best predictor enjoys a closed-form minimum MSE. It provides valuable information about how much the prediction results can be improved. In particular, due to the existence of randomness, the minimum MSE is much greater than $0$. 

Second, the best predictor is indeed a complicated function of $E[\alpha_i(t)|\mathcal{F}_{t-1}]$, $E[\beta_i(t)|\mathcal{F}_{t-1}]$, $E[u_{M,i}(t)|\mathcal{F}_{t-1}]$, $E[v_{M,i}(t)|\mathcal{F}_{t-1}]$, $E[\gamma_i(t)|\mathcal{F}_{t-1}]$, $E[u_i(t)|\mathcal{F}_{t-1}]$ and $E[v_i(t)|\mathcal{F}_{t-1}]$ for $i = 1, 2$. Note that, under the AR processes (\ref{eq:simulation_ar}), 
\begin{equation}
\label{eq:marginal_conditional_expectation}
\begin{aligned}
E[p(t)|\mathcal{F}_{t-1}]  =& \mu_{p} + 0.9p(t-1) - 0.8p(t-2) +\\
 &0.7p(t-3) - 0.6p(t-4) + 0.5p(t-5). 
\end{aligned}
\end{equation}
As a result, a good prediction requires the neural network to pick up the past dependence in the individual series, as well as the function to combine the marginal information. 

\subsection{Experiment Settings and Results}
To make predictions, we include observations of the past 5 steps in neural networks. Apart from $y_i$, parameter series ($\alpha_i, \beta_i$ and so on) are also fed into neural networks, leading the total input dimension $M$ to be 16. 

We test two intuitive ways to group the variables. The first way is to split the variables to two groups, each of which contains $y_i$ together with its parameters, i.e. $m_1 = m_2 = 8$. Since each $y_i$ is jointly determined by its parameters, it may not be wise to break the connections. The second grouping is suggested by the best predictor. The variables are totally split into 16 groups so that there is a marginal component to learn each marginal conditional expectation (\ref{eq:marginal_conditional_expectation}).

\begin{table}
	\centering
	\begin{tabular}{l l l}
		\toprule
		\multirow{3}{*}{Model} & \multirow{3}{*}{MSE} & Relative difference\\
		&&with\\
		&&theoretical minimum\\\midrule
		LSTM&22.26&2.87\%\\
		GRU&22.20&2.59\%\\\midrule
		Channel-wise LSTM & \multirow{2}{*}{22.06}&\multirow{2}{*}{1.94\%}\\
		(two groups)&&\\
		Channel-wise LSTM & \multirow{2}{*}{21.93}&\multirow{2}{*}{1.34\%}\\
		(total split)&&\\\midrule
		mGRN (two groups)&21.91&1.25\%\\
		mGRN (total split)&\textbf{21.88}&\textbf{1.11\%}\\\midrule
		Theoretical minimum & 21.64 &-\\ \bottomrule
	\end{tabular}
	\caption{The average mean squared error (MSE) in the base case of the simulation experiment. The average is taken across the simulation paths generated from 10 pairs of parameters reported in \citet{Yan2019}. The theoretical minimum MSE is calculated with the best predictor $E[100y_1(t)y_2(t)|\mathcal{F}_{t-1}]$. Smaller MSEs indicate better predictions. A relative difference is calculated as (model MSE - theoretical minimum)/theoretical minimum. The bold numbers are the best results. }
	\label{tb:simulation}
\end{table}

A total number of $100,000$ observations are generated for each simulation path. The first $70,000$ observations are used to train models. The next $15,000$ observations are used for validation. The final performance is evaluated in the last $15,000$ observations. As mentioned earlier, all experiments are repeated with 10 pairs of randomly selected stocks reported in \citet{Yan2019}.

In this simulation experiment, we compare the proposed model with a few recurrent neural networks\footnote{The direct output $h$ of a recurrent network has dimension $N$. As a common approach, a linear layer is added at the end of the recurrent layer to convert the $N$-dimensional vector to the final output. The same remark applies to all experiments in the paper. The number of parameters in the dense layer is not counted in controlling model sizes in the simulation experiments.}, namely, LSTM, GRU, and channel-wise LSTM. An advantage of mGRN relative to channel-wise LSTM is that it uses variables efficiently by setting up fewer gates. To demonstrate the advantage, we limit the total number of trainable parameters to be around 1.8 thousand for all models\footnote{The comparison method is also used in \citet{Chung2014} to compare LSTM and GRU. }. The number is chosen such that further increases of the model sizes do not improve validation results for LSTM or GRU. For both channel-wise LSTM and mGRN, we tune $\lambda$ (as mentioned earlier, we set $N=\lambda\tilde{N}$) via grid searches within $\{1, 2, 4, 8\}$. The marginal and joint dimensions ($\tilde{N}$ and $N$) are adjusted accordingly\footnote{Please refer to Appendices for the list of hyperparameters.}. Note that it is impossible to set the numbers of trainable parameters to be exactly the same for all models. In general, we choose the number of parameters of mGRN to be smaller than those of alternative models. We also tune the learning rates via grid searches within $\{10^{-4}, 5\times10^{-4}, 10^{-3}\}$.

\begin{figure}
	\centering
	\includegraphics[width=1\linewidth]{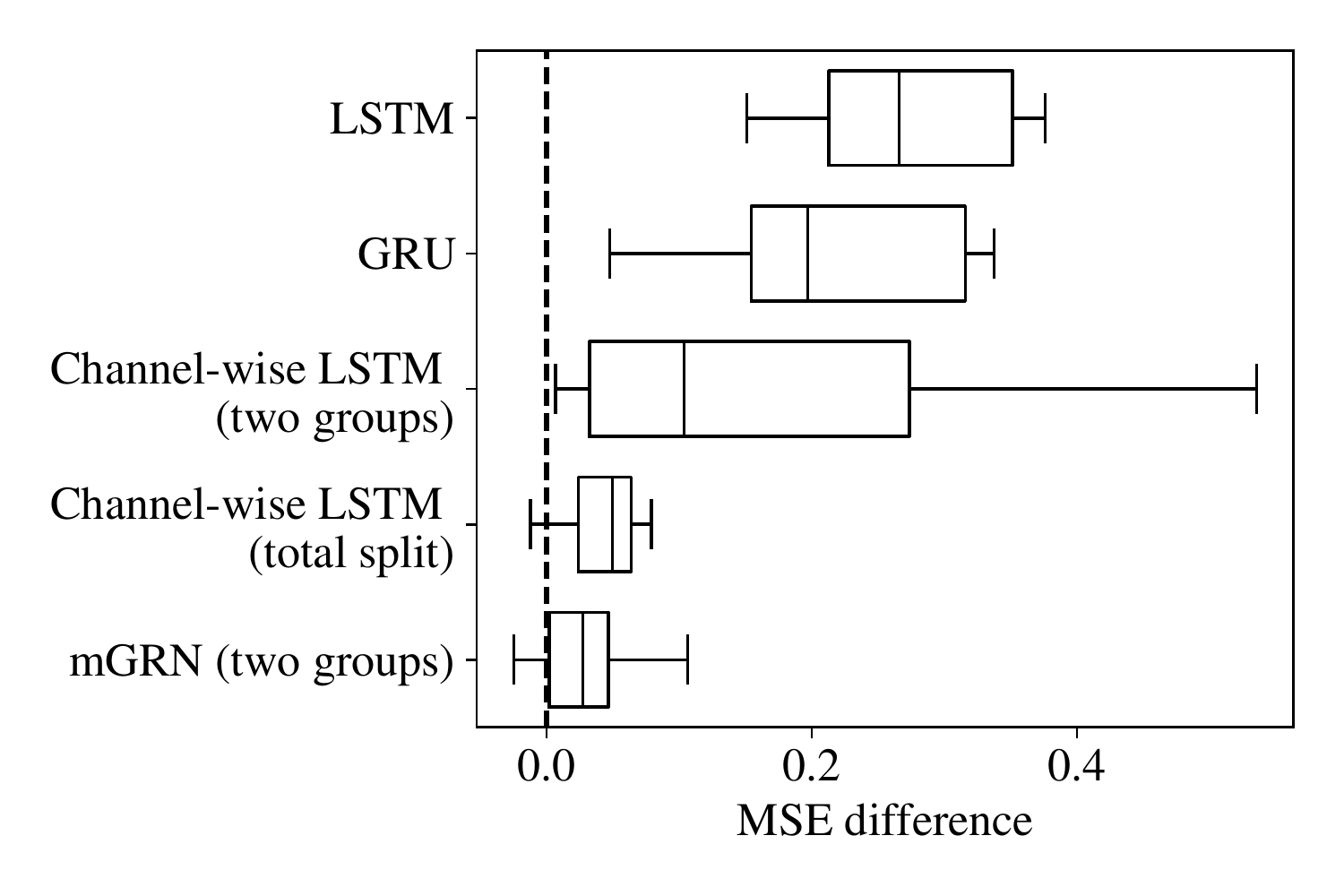}
	\caption{Improvements of mGRN (total split) in MSE comparing to alternative architectures in the simulation experiment. The boxplots are plotted with data simulated from 10 pairs of stock parameters reported in \citet{Yan2019}. The vertical dotted line indicates 0. A positive difference suggests that mGRN (total split) performs better than the corresponding model.  }
	\label{fig:simulation}
\end{figure}

\begin{table*}
	\begin{subtable}[t]{0.47\textwidth}
		\centering
		\begin{tabular}{l l l}
			\toprule
			&AUC-ROC& AUC-PR\\\midrule
			Logistic regression& 0.848 & 0.474\\
			LSTM &0.855&0.485\\
			Channel-wise LSTM &\textbf{0.862}&0.515\\
			mGRN &\textbf{0.862}&\textbf{0.523}\\\bottomrule
		\end{tabular}
		\caption{In-hospital mortality}
	\end{subtable}
	\hspace{2em}
	\begin{subtable}[t]{0.47\textwidth}
		\centering
		\begin{tabular}{l l l}
			\toprule
			&AUC-ROC& AUC-PR\\\midrule
			Logistic regression& 0.870 & 0.214\\
			LSTM &0.892&0.324\\
			Channel-wise LSTM &0.906&0.333\\
			mGRN &\textbf{0.911}&\textbf{0.347}\\
			\bottomrule
		\end{tabular}
		\caption{Decompensation}
	\end{subtable}
	
	\begin{subtable}[t]{0.47\textwidth}
		\centering
		
		\begin{tabular}{l l l}
			\toprule
			&Kappa& MAD\\\midrule
			Logistic regression& 0.402 & 162.3\\
			LSTM &0.438&\textbf{123.1}\\
			Channel-wise LSTM &0.442&136.6\\
			mGRN &\textbf{0.447}&124.6\\
			\bottomrule
		\end{tabular}
		\caption{Length of stay}
	\end{subtable}
	\hspace{2em}
	\begin{subtable}[t]{0.47\textwidth}
		\centering
		\begin{tabular}{l l l}
			\toprule
			&Macro& Micro\\
			&AUC-ROC& AUC-ROC\\\midrule
			Logistic regression& 0.739 & 0.799\\
			LSTM &0.770&0.821\\
			Channel-wise LSTM &0.776&0.825\\
			mGRN &\textbf{0.779}&\textbf{0.826}\\
			\bottomrule
		\end{tabular}
		\caption{Phenotyping}
	\end{subtable}
	\caption{Model performance on the MIMIC-III data set \citep{Johnson2016}. Except for those of mGRN, all results are taken from \citet{Harutyunyan2019}. Greater values are better for all metrics except mean absolute difference (MAD). The bold numbers are the best results. Following \citet{Harutyunyan2019}, the reported results are the mean values calculated by resampling the test sets  $Q$ times with replacement ($Q = 10000$ for in-hospital mortality prediction and phenotype classification, and $Q = 1000$ for decompensation and length-of-stay prediction tasks). $95\%$ confidence intervals are provided in Appendices.}
	\label{tb:mimic_3}
\end{table*}

Table \ref{tb:simulation} gives the average MSE of each model across the 10 simulation paths. We want to emphasize that the improvements of mGRN are not only in the average sense; it outperforms alternative models in almost all simulation cases, as shown in figure \ref{fig:simulation}, which gives the MSE improvements of mGRN (total split). 

There are a few observations. First, both mGRN and channel-wise LSTM perform significantly better than vanilla LSTM and GRU. This is not surprising, as mGRN and channel-wise LSTM are designed for the task. It is more interesting to notice that mGRN performs consistently better than channel-wise LSTM. This suggests that mGRN performs better at picking up the marginal and joint serial dependence. 

A second observation is that predictions are better when variables are totally split. This is consistent with the structure of the best predictor. However, without the best predictor, the result may not be intuitive, as it seems to be unwise to break the tight connections between $y_i$ and its parameters. Of course, there will not be a best predictor to suggest variable grouping. Once again, this may be regarded as a step of hyperparameter tuning, and to totally split the variables may be a good first attempt. 

\section{Real-world Applications}
\label{sec:realdatasets}
In this section, we evaluate the performance of mGRN on real-world data sets. The data sets, together with the results of the state-of-the-art models', are borrowed from \citet{Harutyunyan2019} and \citet{Zhang2020}. All data sets are publicly available. All results, except for those of mGRN, are taken from the two papers. To obtain the results of mGRN, we strictly follow the original experiment settings and perform grid searches on hyperparameters such as variable grouping, dimensions of the marginal and joint components ($\tilde{N}$ and $N$), learning rates, and dropouts. Experiments are coded with Pytorch \citep{Paszke2019} and performed on NVIDIA TITAN Xp GPUs with 12 GB memory. Please refer to Appendices for the codes and pre-trained models to reproduce the results. 

\subsection{MIMIC-III Data Set}
\label{sec:mimic_3}
\citet{Harutyunyan2019} reports the performance of logistic regression, LSTM, and channel-wise LSTM on the Medical Information Mart for Intensive Care (MIMIC-III) data set \citep{Johnson2016}, which consists of multivariate time series of intensive care unit (ICU) records. Raised from real-world applications, the data set contains common difficulties such as missing values and highly skewed responses. 

In their experiments, \citet{Harutyunyan2019} uses 17 clinical variables, including both continuous and categorical types. Categorical variables are encoded in one-hot format. Moreover, each variable is coupled with a binary variable to indicate whether the corresponding variable is observed or not. In total, the input dimension $M$ is 76. Following the variable grouping of channel-wise LSTM, we group each clinical variable with the corresponding indicative binary variable, leading to 17 variable groups.

\begin{table*}
	\centering
	\begin{tabular}{lllllllllll}
		\toprule
		&\multirow{2}{*}{mGRN}&\multirow{2}{*}{TapNet}&MLSTM&WEASEL&ED&DTW&DTW&ED-1NN&DTW-1NN&DTW-1NN\\
		&&&-FCN&+MUSE&-1NN&-1NN-I&-1NN-D&(norm) &-I (norm)&-D (norm)\\\midrule
		CT & 0.995* & \textbf{0.997} & 0.985 & 0.990  & 0.964 & 0.969 & 0.990  & 0.964 & 0.969 & 0.989 \\
		FD  & \textbf{0.596 }      & 0.556* & 0.545 & 0.545 & 0.519 & 0.513 & 0.529 & 0.519 & 0.500 & 0.529 \\
		LSST & \textbf{0.614}                & 0.568 & 0.373 & 0.590* & 0.456 & 0.575 & 0.551 & 0.456 & 0.575 & 0.551 \\
		PD  & \textbf{0.989 }          & 0.980* & 0.978 & 0.948 & 0.973 & 0.939 & 0.977 & 0.973 & 0.939 & 0.977 \\
		PS     & \textbf{0.227}            & 0.175 & 0.110 & 0.190* & 0.104 & 0.151 & 0.151 & 0.104 & 0.151 & 0.151 \\
		SAD   &\textbf{0.997} & 0.983 & 0.990* & 0.982 & 0.967 & 0.960 & 0.963 & 0.967 & 0.959 & 0.963\\
		\bottomrule
	\end{tabular}
	\caption{Classification accuracy on UEA data sets \citep{Bagnall2018}. We focus on the data sets whose training samples are greater than $1000$. They are character trajectories (CT), face detection (FD),  LSST, pen digits (PD), phoneme spectra (PS), and spoken Arabic digits (SAD). Except for those of mGRN, all results are taken from \citet{Zhang2020}. Greater values indicate better classification. The bold numbers are the best results. The second best results are labeled by asterisks. }
	\label{tb:uea}
\end{table*}

\citet{Harutyunyan2019} proposes the following four benchmark health care problems:
\begin{enumerate}
	\item In-hospital-mortality prediction. To predict in-hospital mortality marking use of ICU records in the first 48 hours. This is a binary classification task. 
	\item Decompensation prediction. To predict the mortality of a patient in the next 24 hours making use of all the available ICU records. This is a binary classification task. 
	\item Length-of-stay prediction. To predict the remaining number of days of a patient in the ICU making use of all the available ICU records. The task is a multi-category classification task. 
	\item Phenotype classification. To classify which of 25 acute care conditions are present given the full ICU records. This task is a combination of 25 binary classification tasks. 
\end{enumerate}

As given in table \ref{tb:mimic_3}, classification results are evaluated with multiple metrics, such as the area under the receiver operator
characteristic curve (AUC-ROC) and area under the precision-recall curve (AUC-PR), since the labels are highly skewed \citep{Davis2006}. The total number of samples greatly varies among tasks, ranging from 18 thousand to 3 million. We follow the default training-validation-test split. For more details about the data set and experiment setup, please refer to \citet{Harutyunyan2019}.

Experiment results are reported in table \ref{tb:mimic_3}. Except for those of mGRN, all results are taken from \citet{Harutyunyan2019}. mGRN demonstrates consistent improvements compared with the strong baselines. In particular, benefiting from the deliberately designed gates and information flows, mGRN outperforms channel-wise LSTM even though the two models share a similar intuition. The improvements are the most significant in decompensation prediction and length-of-stay prediction. These two tasks have much greater training samples (around 2 million samples) than the other two tasks (less than 25 thousand samples). Note that, as mentioned earlier, channel-wise LSTM may have gained from doubling the inputs by reversing them, which we purposely exclude from mGRN to avoid confusion.

\subsection{UEA Data Sets}

\citet{Zhang2020} conducts experiments on UEA multivariate time series data sets \citep{Bagnall2018}.  We omit the data sets whose training samples are small and focus on those whose training samples are greater than $1000$\footnote{We exclude the insect wing beat dataset. \citet{Zhang2020} reports the data set has 78 steps, but the version we downloaded from UEA only has 30 steps.}. They are character trajectories (CT), face detection (FD),  LSST, pen digits (PD), phoneme spectra (PS), and spoken Arabic digits (SAD). In these data sets, the number of variables ranges from 2 to 144, and the time series length ranges from 8 to 217. All tasks are multivariate classification evaluated with classification accuracy. Following \citet{Zhang2020}, the data sets are only split to training sets and test sets. 

The results reported in \citet{Zhang2020} comprise a few state-of-the-art algorithms targeting multivariate time series. The algorithms are either extensions of traditional machine learning techniques (ED-I, DTW-I, DTW-D, and WEASEL+MUSE) or novel neural networks (MLSTM-FCN and TapNet). In particular, ED-I, DTW-I, and DTW-D are applied to the data sets with (as labeled by "norm") or without normalization. Please refer to \citet{Zhang2020} for more information about the data sets and experiment setup.


Experiment results are reported in table \ref{tb:uea}.  Except for those of mGRN, all results are taken from \citet{Zhang2020}\footnote{In the published version, \citet{Zhang2020} gives results of 15 data sets. Full results of 30 data sets can be found in the online-companion of \citet{Zhang2020}. }. mGRN dramatically improves classification accuracy on almost all data sets. The greatest improvement is found on the face detection data set. Compared to the second-best result (TapNet), mGRN increases the classification accuracy by $4\%$. The only exception is the character trajectories data set, in which the room to improve is tiny. 

\section{Conclusion}
\label{sec:conclusion}
This paper presents an RNN architecture called Memory-Gated Recurrent Network (mGRN) for the multivariate time series analysis. Motivated by traditional time series models, we introduce separate gates to regulate marginal and joint memories of variables. To be specific, for each variable group, we set up a marginal-memory component in the form of GRU to extract group-specific dependence. Each group's candidate memory is then combined in the joint-memory component to capture interactions among variable groups. Through a combination of comprehensive simulation studies and empirical experiments on a range of public datasets, we show that our proposed mGRN architecture consistently outperforms state-of-the-art architectures targeting multivariate time series. To further improve the architecture performance by combining with other techniques such as CNN and the attention mechanism will be our future research. 
\section*{Acknowledgments}
Qi WU acknowledges the support from the  JD Digits - CityU Joint Laboratory in Financial Technology and Engineering, the Laboratory for AI Powered Financial Technologies, and the  GRF support from the Hong Kong  Research  Grants Council under GRF 14206117 and 11219420. 

Min DAI acknowledges support from Singapore AcRF grants (Grant No. R-703-000-032-112, R-146-000-306-114 and R-146-000-311-114), and the National Natural Science Foundation of China (Grant 11671292).
\bibliography{aaai_submission}
\end{document}


\maketitle
\section{Simulation experiments}
To fix the constant terms in the data generation processes, 10 stock pairs are randomly selected from Table 2 of \citet{Yan2019}. The stock pairs are (IBM, KO),	 (BA, CAT),	 (DWDP, JNJ),	 (CVX, PG),	 (IBM, JNJ),	 (NKE, WMT),	 (BA, PG),	 (INTC, KO),	 (AAPL, NKE),	 (MMM, DIS). The constant terms corresponding to each stock are listed in Table \ref{tb:simulation_constants}.

\begin{table}
	\centering
	\begin{tabular}{llllllll}
		\toprule
		& $\mu_\alpha$&$\mu_{\log\beta}$&$\mu_{\log u_M}$&$\mu_{\log v_M}$&$\mu_{\log\gamma}$&$\mu_{\log u}$&$\mu_{\log v}$\\\midrule
		AAPL & 0.008  & -1.024 & 0.000 & 0.175 & -0.840 & 0.215 & 0.159  \\
		BA   & -0.007 & -1.026 & 0.183 & 0.182 & -0.842 & 0.164 & 0.120  \\
		CAT  & 0.020  & -0.975 & 0.000 & 0.202 & -0.847 & 0.199 & 0.153\\
		CVX  & 0.011  & -1.021 & 0.000 & 0.193 & -0.849 & 0.172 & 0.138  \\
		DIS  & 0.002  & -1.001 & 0.156 & 0.214 & -0.862 & 0.196 & 0.151  \\
		DWDP & -0.007 & -0.994 & 0.176 & 0.186 & -0.866 & 0.198 & 0.141  \\
		IBM  & 0.021  & -0.942 & 0.000 & 0.198 & -0.886 & 0.218 & 0.178  \\
		INTC & 0.012  & -0.948 & 0.000 & 0.149 & -0.873 & 0.168 & 0.141 \\
		JNJ  & -0.003 & -1.012 & 0.189 & 0.210 & -0.858 & 0.227 & 0.160 \\
		KO   & 0.007  & -0.979 & 0.117 & 0.198 & -0.856 & 0.208 & 0.153  \\
		MMM  & 0.001  & -0.964 & 0.186 & 0.198 & -0.862 & 0.199 & 0.161  \\
		NKE  & -0.002 & -0.995 & 0.267 & 0.200 & -0.793 & 0.347 & 0.297 \\
		PG   & 0.010  & -0.979 & 0.096 & 0.201 & -0.844 & 0.210 & 0.161  \\
		WMT  & -0.007 & -0.984 & 0.183 & 0.142 & -0.871 & 0.181 & 0.146 \\
		\bottomrule 
	\end{tabular}
\caption{Constant terms of parameter processes matching with randomly selected stocks from table 2 of \citet{Yan2019}. }
\label{tb:simulation_constants}
\end{table}
%

%

Component dimensions of neural networks in the simulation experiment are given in table \ref{tb:n_units}. 

\begin{table}
	\centering
	\begin{tabular}{lllll}
		\toprule
		&$\lambda$&$\tilde{N}$&$N$&Number of trainable parameters\\\midrule
		LSTM&-&-&14&1736\\\hline
		GRU&-&-&17&1734\\\hline
		\multirow{4}{*}{Channel-wise LSTM (two groups)}&1&5&5&1640\\
		&2&4&8&1632\\
		&4&3&12&1776\\
		&8&2&16&1952\\\hline
		\multirow{4}{*}{Channel-wise LSTM (total split)}&1&3&3&3120\\
		&2&2&4&2128\\
		&4&2&8&3360\\
		&8&2&16&6208\\\hline
		\multirow{4}{*}{mGRN (two groups)}&1&10&10&1620\\
		&2&8&16&1616\\
		&4&6&24&1836\\
		&8&3&24&1368\\\hline
		\multirow{4}{*}{mGRN (total split)}&1&4&4&1496\\
		&2&4&8&1872\\
		&4&3&12&1656\\
		&8&2&16&1440\\ \bottomrule
	\end{tabular}
	\caption{Component dimensions and number of trainable parameters of neural networks in the simulation experiment. We limit the total number of trainable parameters to be around 1.8 thousand without including the parameters in the output dense layer. The number is chosen such that to further increase model sizes does not improve validation results for LSTM or GRU. It is impossible to set the numbers of trainable parameters to be exactly the same for all models. In general, we choose the number of parameters of mGRN to be smaller than those of alternative models. Moreover, to obtain a reasonable performance from channel-wise LSTM, we set a lower limit of 2 for the marginal component dimension. This indeed leads to much greater models when variables are totally split in channel-wise LSTM.}
	\label{tb:n_units}
\end{table}

\subsection{Theoretical minimum MSE}
When MSE is chosen to be the loss function of the prediction problem, the best predictor is the conditional expectation \citep{Shumway2017}. 
\begin{equation}
\label{eq:product_conditional_expectation}
\begin{aligned}
E[100y_1(t)y_2(t)|\mathcal{F}_{t-1}] =& 100\left(\left(E[\alpha_1(t)|\mathcal{F}_{t-1}]+E[\gamma_1(t)|\mathcal{F}_{t-1}]E[g(\omega_1(t);u_{1}(t), v_{1}(t))|\mathcal{F}_{t-1}]\right)E[y_2(t)|\mathcal{F}_{t-1}]\right.+\\
& E[\beta_1(t)|\mathcal{F}_{t-1}]E[g(\omega_M(t);u_{M,1}(t), v_{M,1}(t))|\mathcal{F}_{t-1}]\times\\
&\left(E[\alpha_2(t)|\mathcal{F}_{t-1}]+ E[\gamma_2(t)|\mathcal{F}_{t-1}]E[g(\omega_2(t);u_{1}(t), v_{2}(t))|\mathcal{F}_{t-1}]\right)+\\
&E[\beta_1(t)|\mathcal{F}_{t-1}] E[\beta_2(t)|\mathcal{F}_{t-1}]\times\\
& \left.E[g(\omega_M(t);u_{M,1}(t), v_{M,1}(t))g(\omega_M(t);u_{M,2}(t), v_{M,2}(t))|\mathcal{F}_{t-1}]\right).
\end{aligned}
\end{equation}

To evaluate equation (\ref{eq:product_conditional_expectation}), a primary observation is that the conditional distribution of an AR process is Gaussian. To be specific, we consider the following AR(5) process
\begin{equation*}
p(t) = \mu + \sum_{j=1}^{5}\zeta_jp(t-j) + \epsilon(t), 
\end{equation*}
with Gaussian random noise $\epsilon(t)\sim N(0, \sigma^2)$. Given observations up to step $t-1$, the conditional distribution of $p(t)$ is Gaussian with mean $\phi_p(t) =  \mu + \sum_{j=1}^{5}\zeta_jp(t-j) $ and variance $\sigma^2$. Moreover,
\begin{equation*}
E[\exp(p(t))|\mathcal{F}_{t-1}] = \exp\left(\phi_p(t) +\frac{\sigma^2}{2} \right).
\end{equation*}
This enables us to evaluate $E[\alpha_i(t)|\mathcal{F}_{t-1}]$, $E[\beta_i(t)|\mathcal{F}_{t-1}]$, $E[u_{M,i}(t)|\mathcal{F}_{t-1}]$, $E[v_{M,i}(t)|\mathcal{F}_{t-1}]$, $E[\gamma_i(t)|\mathcal{F}_{t-1}]$, $E[u_i(t)|\mathcal{F}_{t-1}]$ and $E[v_i(t)|\mathcal{F}_{t-1}]$. 

It remains to evaluate the conditional expectations involving function $g$. Given $\omega\sim N(0,1)$, and  $u(t)$ whose conditional distribution is lognormal with parameters $\phi_{\log u}(t)$ and  $\sigma_{\log u}^2<1$, we define
\begin{equation*}
\begin{aligned}
&V_1(u(t)) := E\left[\omega u(t)^{\omega}|\mathcal{F}_{t-1}\right] = \frac{\phi_{\log u}(t)}{\left(1-\sigma_{\log u}^2\right)^{3/2}}\exp\left(\frac{\phi_{\log u}^2(t)}{2-2\sigma_{\log u}^2}\right),\\
&V_2(u(t)) := E\left[\omega^2u(t)^{\omega}|\mathcal{F}_{t-1}\right] = \frac{1 + \phi_{\log u}^2(t) - \sigma_{\log u}^2}{\left(1-\sigma_{\log u}^2\right)^{5/2}}\exp\left(\frac{\phi_{\log u}^2(t)}{2-2\sigma_{\log u}^2}\right).
\end{aligned}
\end{equation*}
The conditional expectations involving function $g$ can be evaluated with $V_1$ and $V_2$. 
\begin{equation}
\label{eq:g_conditional_expectation}
\begin{aligned}
&E[g(\omega_M(t);u_{M,i}(t), v_{M,i}(t)|\mathcal{F}_{t-1}] = \frac{1}{A} \left(V_1(u_{M,i}(t)) +  V_1\left(\frac{1}{v_{M,i}(t)}\right)\right),\\ 
&E[g(\omega_i(t);u_{i}(t), v_{i}(t)|\mathcal{F}_{t-1}] =  \frac{1}{A} \left(V_1(u_{i}(t)) +  V_1\left(\frac{1}{v_{i}(t)}\right)\right),\\
&E[g(\omega_M(t);u_{M,1}(t), v_{M,1}(t))g(\omega_M(t);u_{M,2}(t), v_{M,2}(t))|\mathcal{F}_{t-1}] \\
= & \frac{1}{A^2}\left(V_2(u_{M,1}(t)u_{M,2}(t)) + V_2\left(\frac{u_{M,1}(t)}{v_{M,2}(t)}\right) + V_2\left(\frac{u_{M,2}(t)}{v_{M,1}(t)}\right) + V_2\left(\frac{1}{v_{M,1}(t)v_{M,2}(t)}\right) \right)+\\
&\frac{1}{A}\left(V_2(u_{M,1}(t)) +V_2(u_{M,2}(t)) +V_2\left(\frac{1}{v_{M,1}(t)}\right) +V_2\left(\frac{1}{v_{M,2}(t)}\right) \right) + 1.
\end{aligned}
\end{equation}

Lastly, by substituting the respective parts into equation (\ref{eq:product_conditional_expectation}), we have
\begin{equation*}
\begin{aligned}
E[100y_1(t)y_2(t)|\mathcal{F}_{t-1}] =& 100\left(\phi_{\alpha_1(t)}E[y_2(t)|\mathcal{F}_{t-1}]\right.+\\
&\exp\left(\phi_{\log{\gamma_1}} + \frac{\sigma_{\log{\gamma_1}}^2}{2}\right)E[g(\omega_1(t);u_{1}(t), v_{1}(t))|\mathcal{F}_{t-1}]E[y_2(t)|\mathcal{F}_{t-1}]+ \\
& \exp\left(\phi_{\log{\beta_1}} + \frac{\sigma_{\log{\beta_1}}^2}{2}\right)E[g(\omega_M(t);u_{M,1}(t), v_{M,1}(t))|\mathcal{F}_{t-1}]\times\\
&\left(\phi_{\alpha_2(t)}+ \exp\left(\phi_{\log{\gamma_2}} + \frac{\sigma_{\log{\gamma_2}}^2}{2}\right)E[g(\omega_2(t);u_{1}(t), v_{2}(t))|\mathcal{F}_{t-1}]\right)+\\
&\exp\left(\phi_{\log{\beta_1}} + \frac{\sigma_{\log{\beta_1}}^2}{2}\right) \exp\left(\phi_{\log{\beta_2}} + \frac{\sigma_{\log{\beta_2}}^2}{2}\right)\times\\
& \left.E[g(\omega_M(t);u_{M,1}(t), v_{M,1}(t))g(\omega_M(t);u_{M,2}(t), v_{M,2}(t))|\mathcal{F}_{t-1}]\right),
\end{aligned}
\end{equation*}
where
\begin{equation*}
\begin{aligned}
E[y_2(t)|\mathcal{F}_{t-1}] =& \phi_{\alpha_2(t)} + \exp\left(\phi_{\log{\beta_2}} + \frac{\sigma_{\log{\beta_2}}^2}{2}\right)E[g(\omega_M(t);u_{M,2}(t), v_{M,2}(t))|\mathcal{F}_{t-1}] + \\
&\exp\left(\phi_{\log{\gamma_2}} + \frac{\sigma_{\log{\gamma_2}}^2}{2}\right)E[g(\omega_2(t);u_{1}(t), v_{2}(t))|\mathcal{F}_{t-1}],\\
\phi_p(t) =& \mu_{p} + 0.9p(t-1) - 0.8p(t-2) + 0.7p(t-3) - 0.6p(t-4) + 0.5p(t-5)\\
& \text{for}\quad  p = \alpha_i, \log{\beta_i}, \log{u_{M, i}}, \log{v_{M, i}}, \log{\gamma_i}, \log{u_i}, \log{v_i} \quad \text{and} \quad i = 1, 2,\\
\sigma_p =& 0.1\quad \text{for}\quad  p = \alpha_i, \log{\beta_i}, \log{u_{M, i}}, \log{v_{M, i}}, \log{\gamma_i}, \log{u_i}, \log{v_i} \quad \text{and} \quad i = 1, 2,\\
\end{aligned}
\end{equation*}
and conditional expectations involving function $g$ are given by equation (\ref{eq:g_conditional_expectation}).
\section{MIMIC-III Data Set}
The full experiment results (including the $95\%$ confidence intervals) are given in table \ref{tb:mimic_3}. 

\begin{table}
	\footnotesize
	\begin{subtable}[t]{0.5\textwidth}
		\centering
		\begin{tabular}{l l l}
			\toprule
			&AUC-ROC& AUC-PR\\\midrule
			Logistic& 0.848 & 0.474\\
			 regression&(0.828, 0.868)&(0.419, 0.529)\\\midrule
			LSTM &0.855&0.485\\
			&(0.835, 0.873)&(0.431, 0.537)\\\midrule
			Channel-wise &\textbf{0.862}&0.515\\
			LSTM&(0.844, 0.881)&(0.464, 0.568)\\\midrule
			mGRN &\textbf{0.862}&\textbf{0.523}\\
			&(0.843, 0.880)&(0.469, 0.575)\\\bottomrule
		\end{tabular}
		\caption{In-hospital mortality}
	\end{subtable}
	\hspace{1em}
	\begin{subtable}[t]{0.5\textwidth}
		\centering
		\begin{tabular}{l l l}
			\toprule
			&AUC-ROC& AUC-PR\\\midrule
			Logistic & 0.870 & 0.214\\
			regression&(0.867, 0.873)&(0.205, 0.223)\\\midrule
			LSTM &0.892&0.324\\
			&(0.889, 0.895)&(0.314, 0.333)\\\midrule
			Channel-wise  &0.906&0.333\\
			LSTM&(0.903, 0.909)&(0.323, 0.344)\\\midrule
			mGRN &\textbf{0.911}&\textbf{0.347}\\
			&(0.908, 0.913)&(0.338, 0.358)\\
			\bottomrule
		\end{tabular}
		\caption{Decompensation}
	\end{subtable}
	
	\begin{subtable}[t]{0.5\textwidth}
		\centering
		\begin{tabular}{l l l}
			\toprule
			&Kappa& MAD\\\midrule
			Logistic & 0.402 & 162.3\\
			regression&(0.401, 0.404)&(161.8, 162.8)\\\midrule
			LSTM &0.438&\textbf{123.1}\\
			&(0.436, 0.440)&(122.6, 123.5)\\\midrule
			Channel-wise &0.442&136.6\\
			LSTM&(0.440, 0.444)&(136.1, 137.1)\\\midrule
			mGRN &\textbf{0.447}&124.6\\
			&(0.445, 0.449)&(124.1, 125.0)\\
			\bottomrule
		\end{tabular}
		\caption{Length of stay}
	\end{subtable}
	\hspace{1em}
	\begin{subtable}[t]{0.5\textwidth}
		\centering
		\begin{tabular}{l l l}
			\toprule
			&Macro& Micro\\
			&AUC-ROC& AUC-ROC\\\midrule
			Logistic& 0.739 & 0.799\\
			regression&(0.734, 0.743)&(0.796, 0.803)\\\midrule
			LSTM &0.770&0.821\\
			&(0.766, 0.775)&(0.818, 0.825)\\\midrule
			Channel-wise &0.776&0.825\\
			LSTM&(0.772, 0.781)&(0.822, 0.828)\\\midrule
			mGRN &\textbf{0.779}&\textbf{0.826}\\
			&(0.774, 0.783)&(0.823, 0.830)\\
			\bottomrule
		\end{tabular}
		\caption{Phenotyping}
	\end{subtable}
	\caption{Model performance on the MIMIC-III data set \citep{Johnson2016}. Except for those of mGRN, all results are taken from \citet{Harutyunyan2019}. Greater values are better for all metrics except mean absolute difference (MAD). The bold numbers are the best results. Following \citet{Harutyunyan2019}, the reported results are the mean values calculated by resampling the test sets  $Q$ times with replacement ($Q = 10000$ for in-hospital mortality prediction and phenotype classification, and $Q = 1000$ for decompensation and length-of-stay prediction tasks). $95\%$ confidence intervals are given in parentheses.}
	\label{tb:mimic_3}
\end{table}

\newpage
\bibliographystyle{apalike}
\bibliography{supplement}